\documentclass[a4paper,fleqn]{cas-dc}

\usepackage[authoryear,longnamesfirst]{natbib}
\usepackage{graphicx}
\usepackage{algorithm}
\usepackage{algorithmic}
\usepackage{amsmath}
\usepackage{amssymb}
\usepackage{etoolbox}
\usepackage{xspace}

\def\tsc#1{\csdef{#1}{\textsc{\lowercase{#1}}\xspace}}
\tsc{WGM}
\tsc{QE}

\begin{document}
\let\WriteBookmarks\relax
\def\floatpagepagefraction{1}
\def\textpagefraction{.001}

\shorttitle{Optimal Path Planning with Localization 
constraints}    

\shortauthors{Mundane et al.}  

\title [mode = title]{Optimal Path Planning with Localization 
constraints}  



\author[1]{Chinmay Mundane}\cormark[1]
\ead{chinmaymundane.123@gmail.com}

\author[1]{Manav Mishra}
\ead{mishra20@iiserb.ac.in}

\author[1]{Sujit PB}
\ead{sujit@iiserb.ac.in}

\affiliation[1]{organization={IISER Bhopal},
            country={India}}

%













\cortext[1]{Corresponding author}



\begin{abstract}
This paper studies UAV path planning in areas where GPS signals are not present. Because these drones can easily get lost when they don't have GPS, they must rely on dead reckoning and occasionally pass near cellular towers, which act as landmarks to reset and fix their position. The goal is to find the shortest path from a start to a goal location while keeping the estimation uncertainty within a given bound. To solve this problem, we propose a localization-constrained A* planner that treats uncertainty as a hard constraint during search. The robot state and covariance are propagated using an Extended Kalman Filter (EKF), and only paths that satisfy the covariance limit are considered. We also show that the proposed method does not affect heuristic used in A* and remains admissible and consistent, ensuring optimality among all 
feasible paths. 
Extensive Monte Carlo simulations and PyBullet quadrotor test show that the proposed method finds shorter and more reliable paths compared to RRT* and PRM, while always satisfying the uncertainty constraint.
\end{abstract}





\maketitle


\section{INTRODUCTION}

The use of autonomous unmanned aerial vehicles (UAVs) has expanded rapidly, driven by affordable platforms and sensors. Applications ranging from search and rescue operations in disaster zones to autonomous delivery in remote areas often encounter GPS-denied conditions due to signal interference, multipath effects, or insufficient satellite visibility. In such scenarios, robots must navigate using alternative localization methods, typically combining their local sensors, such as inertial measurement units (IMUs) and external aids like observations of nearby landmarks, to localize and navigate without GPS.

\begin{figure}
    \centering
    \includegraphics[width=0.98\columnwidth]{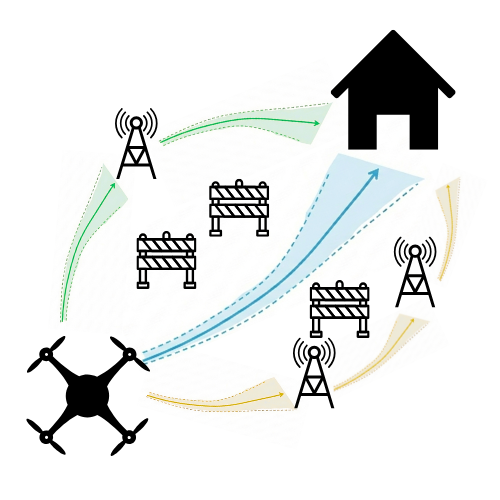}
    \caption{Illustration of the planning problem considered in this work. The robot must navigate from a fixed start to a goal location while avoiding obstacles and periodically using landmarks to bound state uncertainty.}
    \label{fig:problem-introduction}
\end{figure}

The navigation is challenging not only because of obstacle avoidance, but also because localization uncertainty grows as the robot moves without absolute position updates due to sensor noise and external disturbances like wind. Most classical path-planning algorithms focus on geometric optimality measures like distance or time, but they generally assume that the robot’s pose is known exactly or with negligible error like in \cite{c1,c2,c9}. This can lead to paths that are optimal in geometry but pass through regions with poor sensing, causing the estimation uncertainty to increase beyond acceptable limits. When this happens, the robot may be forced to replan or may fail to complete the mission. Therefore, planning methods that explicitly account for localization uncertainty during path generation are essential for reliable operation in GPS-denied settings.

At the same time, UAVs have limited battery life, so they cannot afford to take very long paths. A longer path may help reduce localization error by passing through areas with better sensing, but it also increases flight time and energy use. This creates a trade-off: short paths are energy-efficient but risky for localization, while safer paths may be too long. So, we need to find paths that are both short and keep the localization error within acceptable limits.

In this context, consider a scenario given in figure \ref{fig:problem-introduction}, where a robot needs to reach a goal where GPS may not be available all the time. A geometrically direct path (shown in blue), typical of classical geometry-based planners can get lost and risk mission success due to localization uncertainty (shown in dotted blue). Alternative paths that exploit landmark sensing regions are shown in green and yellow. Sampling-based planners may instead produce a longer landmark-assisted path (shown in yellow) that remains feasible but increases flight time and energy consumption. Our method selects the shortest localization-feasible path (shown in green), while also satisfying the covariance bounds.   \

We address the problem of path planning in GPS-denied environments with sparse localization zones. The environment is assumed to be known, and sensor models are available to predict the effect of motion and observations on the robot’s state estimate. Our goal is to compute paths that explicitly bound state-estimation error while remaining as short as possible, ensuring both localization safety and mission efficiency. 

In the literature, work on cooperative localization like ~\cite{1668252,8049492,7120118,9250516} and path planning~\cite{Song,9037111,9384209} have largely been studied independently. Localization in GPS-denied environments has been a significant challenge in the field of autonomous systems. The early work focused solely on improving navigation estimation accuracy, e.g.,~\cite{ding2007},~\cite{hide2003}. Many of these approaches have often relied on filtering techniques and probabilistic roadmaps to mitigate localization errors. While effective, these methods typically rely on sampling and do not provide optimality guarantees.

A related work ~\cite{c4} studies path planning with localization constraints using a graph built from a probabilistic roadmap (PRM). This helps deal with uncertainty, but since it is based on sampling, it does not guarantee the best optimal solution. The authors improve their method by creating a larger graph to handle multiple objectives. However, their approach assumes the robot can move freely in any direction (holonomic motion). In contrast, our method considers more realistic motion constraints (non-holonomic) by generating smooth, continuous paths between points using a carrot-following controller.

Another related work ~\cite{c3} proposes a time-optimal path planner that keeps the localization uncertainty within limits using Particle Swarm Optimization and a carrot-following controller. They use cellular towers as landmarks for UAV localization. However, the method only finds a near-optimal solution and does not guarantee optimality. In ~\cite{SundarKaarthik2019Ppfu}, the authors take a different approach by placing landmarks optimally in the environment to satisfy localization constraints. Their method computes both the robot’s path and the best locations to place these landmarks. In contrast, our work assumes that landmarks (such as cellular towers) are fixed and sparse, which better reflects real-world urban environments.

Planning under uncertainty has also been addressed through belief-space planning, where the belief over the system state is treated as the planning state. Approaches such as \cite{platt2010belief} formulate navigation in partially observable environments as a POMDP and compute policies over belief dynamics. Their method requires reliable belief updates and explicit belief transitions, which may be violated in GPS-denied settings with intermittent or weak observations. Similarly, \cite{10770054} studies decision-making under uncertainty by formulating navigation and selective localization as a constrained POMDP and learning policies that balance localization frequency against mission risk. This approach, while powerful for when to localize, still relies on maintaining a belief over states and solving a complex POMDP or reinforcement learning problem over that belief, which can be computationally expensive.

In addition to planning-based approaches, another line of work focuses on improving localization through better state estimation techniques. Moving-horizon estimation (MHE) has been suggested as an alternative to EKF to increase the accuracy of nonlinear estimation problems by~\cite{RAWLINGS20061529},~\cite{WANG20141581}, and~\cite{10.1007/s10846-016-0408-2}. For instance ~\cite{8901098} proposed an approach combining NMPC and pose-graph-MHE for 3D formation control of micro aerial vehicles with relative sensing capability. 
The use of Nonlinear Model Predictive Control (NMPC) for cooperative localization in multi-agent systems has also been explored in~\cite{8797888}, where they proposed a joint cooperative path planning algorithm and localization for GPS-denied areas. Their approach combines NMPC for optimal path generation and a Moving Horizon Estimator (MHE) for state estimation, reducing the computational burden of traditional Extended Kalman Filter (EKF) methods. Although effective, NMPC-based methods typically incur high computational costs and lack the deterministic completeness and optimality guarantees provided by search-based planners.

The key contributions of this paper are: \begin{itemize}
    \item We propose a path planning method for robots in GPS-denied environments that uses cellular towers as localization landmarks.
    \item The planner minimizes travel distance while explicitly constraining state-estimation uncertainty arising from dead reckoning.
    \item Non-holonomic A* planning is used to compute feasible paths with A* guarantees.
    \item Comparison across other planners like RRT*, and PRM
\end{itemize}

\section{METHODOLOGY}
In this section, we describe how we model the robot's motion, the environment, and covariance constraint into the planner. Then we also explain the method to implement EKF for A* discretized nodes as a continuous-time dependent system.

\subsection{Motion and Sensing Model}

The robot is assumed to operate in a 2D space and is modeled as a Dubins vehicle moving at constant linear velocity. The robot state at time $t$ is defined as
\begin{equation}
\mathbf{x}(t) =
\begin{bmatrix}
x(t) \\
y(t) \\
\psi(t)
\end{bmatrix},
\end{equation}
where $x(t)$ and $y(t)$ denote the position, and $\psi(t)$ represents the heading angle. The agent is equipped with range and bearing
sensors in order to estimate the state.

The control input is the angular velocity
\begin{equation}
u(t) = \dot{\psi}(t),
\end{equation}
while the linear velocity $v$ remains constant. To account for external disturbances and improve robustness, we explicitly incorporate wind effects into the motion model. Specifically, we modify the standard Dubins vehicle dynamics to include a wind velocity component. The resulting discrete-time kinematic equations, incorporating wind disturbance, are taken from \cite{wind_eq} and given by:

\begin{align}
x_{k+1} &= x_k + (v \cos\psi_k + v_w \cos\psi_w)\Delta t, \\
y_{k+1} &= y_k + (v \sin\psi_k + v_w \sin\psi_w)\Delta t, \\
\psi_{k+1} &= \psi_k + \frac{\omega_k}{\|\mathbf{v}_g\|}\Delta t,
\end{align}
where $v$ is the commanded airspeed, $\psi_k$ is the heading angle,
$v_w$ is the wind speed, $\psi_w$ is the wind direction,
$\omega_k$ is the angular control input, and the ground speed is
$\|\mathbf{v}_g\| = \sqrt{(v\cos\psi_k + v_w\cos\psi_w)^2 +
(v\sin\psi_k + v_w\sin\psi_w)^2}$.

For simulation and estimation, the model is discretized using a sampling time $\Delta t$, yielding
\begin{equation}
\mathbf{x}_{k+1} = \mathbf{f}(\mathbf{x}_k, u_k) + \boldsymbol{\eta}_k,
\end{equation}
where $\boldsymbol{\eta}_k$ is the process noise, modeled as an independently generated zero-mean Gaussian random vector with covariance $Q$.

The environment contains a set of known landmarks with a finite sensing range. When the robot lies within the sensing region of a landmark, it receives a precise measurement related to its state, which is treated as a high-confidence observation. When no landmark is observable, no measurement is available, and the robot relies entirely on dead reckoning.

\subsection{EKF-Based State Estimation}

The robot maintains a Gaussian belief over its state, represented by the estimated mean $\hat{\mathbf{x}}_k$ and covariance $P_k$. State estimation is performed using an Extended Kalman Filter (EKF) to account for the non-linearities in the motion and sensing models.

The prediction step is given by:
\begin{equation}
\hat{\mathbf{x}}_{k+1}^{-} = \mathbf{f}(\hat{\mathbf{x}}_k, u_k),
\end{equation}
\begin{equation}
P_{k+1}^{-} = F_k P_k F_k^\top + Q,
\end{equation}
where $\mathbf{f}$ describes the non-linear discrete-time system map. The matrix $F_k$ is the linearization of $\mathbf{f}$ around the current estimate $(\hat{\mathbf{x}}_k, 0)$, representing the Jacobian of the motion model with respect to the state.

When a measurement is available from a landmark, the update equations are:
\begin{equation}
K_k = P_{k+1}^{-} H_k^\top \left( H_k P_{k+1}^{-} H_k^\top + R \right)^{-1},
\end{equation}
where $H_k$ is the observation matrix that linearizes the relationship between the state and the sensor output around the predicted state $\mathbf{f}(\hat{\mathbf{x}}_k, 0)$. This Jacobian matrix effectively maps the high-dimensional state space into the observation space, enabling the filter to compute the residual and update the belief.

Here, $R$ is the measurement noise covariance used to regulate the Kalman gain $K_k$ and prevent overconfident state estimates. The estimated mean and covariance thus become:
\begin{equation}
\hat{\mathbf{x}}_{k+1} = \hat{\mathbf{x}}_{k+1}^{-} + K_k (\mathbf{y}_k - H_k \hat{\mathbf{x}}_{k+1}^{-}),
\end{equation}
\begin{equation}
P_{k+1} = (I - K_k H_k) P_{k+1}^{-}.
\end{equation}

\subsection{Uncertainty Metric}

To quantify localization uncertainty, we can use any scalar metric derived from the EKF covariance. We use the trace of the positional covariance, which is defined as
\begin{equation}
\Lambda(P_k) = \mathrm{Tr}(P_{k}),
\end{equation}

We use the trace of the covariance matrix because it is symmetric and positive semi-definite, hence all diagonal entries are non-negative, giving us a positive scalar quantity.
\begin{equation}
\Lambda(P_k) \ge 0 \quad \forall k.
\end{equation}

and thus formulating it as a constraint:
\begin{equation}
\Lambda(P_k) \le \Sigma_{\max},
\end{equation}
where $\Sigma_{\max}$ is a user-defined threshold. Any trajectory segment between two nodes that violates this constraint is considered infeasible.

\subsection{Path-Following Control}

To execute and evaluate a feasible candidate path segment between 2 nodes, a path-following controller is used to generate the control input $u$. The control u can be determined using any path following algorithms. In this paper, we use the basic path following algorithm described in \cite{pathfollow} – the rabbit-carrot based path following.
The controller defines a look-ahead point along the path segment and steers the robot toward this point by generating the angular velocity command $u$, while maintaining a constant forward velocity, which can be seen in Algorithm~\ref{alg:controller}.

\begin{algorithm}[h]
\caption{Path-Following Controller}
\label{alg:controller}
\begin{algorithmic}[1]

\STATE \textbf{Initialize:}
Waypoint $W_i=(x_i,y_i)$, next waypoint $W_{i+1}=(x_{i+1},y_{i+1})$, 
robot pose $p=(x,y)$, heading $\psi$, lookahead $\delta$

\STATE Compute desired direction:
$\theta_d = \operatorname{atan2}(y_{i+1}-y_i,\; x_{i+1}-x_i)$

\STATE Compute robot-to-waypoint angle:
$\theta_r = \operatorname{atan2}(y - y_i,\; x - x_i)$

\STATE Heading difference:
$\beta = \theta_d - \theta_r$

\STATE Distance to waypoint:
$R_u = \|W_{i+1} - p\|$

\STATE Projection distance:
$R = \sqrt{R_u^2 - (R_u \sin\beta)^2}$

\STATE Lookahead (carrot) point:
$(x_t',y_t') = ((R+\delta)\cos\theta_d,\; (R+\delta)\sin\theta_d)$

\STATE Desired heading:
$\psi_d = \operatorname{atan2}(y_t'-y,\; x_t'-x)$

\STATE Control law:
$u = \kappa(\psi_d - \psi)$

\end{algorithmic}
\end{algorithm}

\subsection{Uncertainty-Aware Edge Evaluation}

After a segment is created  using Algorithm~\ref{alg:controller}, each candidate path segment is evaluated using the uncertainty-check function, summarized in Algorithm~\ref{alg:validate}. The robot motion and EKF propagation are simulated simultaneously, and the uncertainty metric is monitored throughout execution, rejecting the node that does not follow the constraint.

\begin{algorithm}[h]
\caption{Uncertainty-Check Function}
\label{alg:validate}
\begin{algorithmic}[1]
\STATE Initialize state, EKF mean, and covariance
\WHILE{target node not reached}
    \STATE Compute control input using Algorithm~\ref{alg:controller}
    \STATE Propagate robot state
    \STATE Perform EKF prediction
    \IF{measurement available}
        \STATE Perform EKF update
    \ENDIF
    \IF{$\Lambda(P_k) > \Sigma_{\max}$}
        \STATE Reject target node and terminate
    \ENDIF
\ENDWHILE
\STATE Accept node
\end{algorithmic}
\end{algorithm}

\subsection{Uncertainty-Constrained A* Planning}

Path planning is performed using an A* search on a discrete graph. The algorithm maintains an OPEN set of candidate nodes and a CLOSED set of expanded nodes. Each node is associated with a cost-to-come $g(\cdot)$, a heuristic $h(\cdot)$, and an evaluation function $f(\cdot)=g(\cdot)+h(\cdot)$.

At each iteration, the node $n_{\text{best}}$ with minimum $f$ is selected from the OPEN set and moved to the CLOSED set. If $n_{\text{best}}$ corresponds to the goal, the search terminates and the path is recovered using backpointers.

For expansion, all successors $x \in \mathrm{Star}(n_{\text{best}})$ that are not in the CLOSED set are considered. Each transition $(n_{\text{best}},x)$ is first validated using an uncertainty-check function. The robot motion is simulated using the Dubins model and a path-following controller, while the belief state is propagated using an Extended Kalman Filter. The trace of the positional covariance is monitored throughout the simulation, and the transition is rejected if the uncertainty exceeds a prescribed threshold.

Only successors that satisfy the uncertainty constraint are added to the OPEN set. Standard A* update rules are then applied to initialize or update the cost-to-come and backpointers. In this way, the search explores only dynamically feasible and uncertainty-safe transitions, yielding the shortest path that satisfies the localization uncertainty constraint.

\begin{algorithm}[h]
\caption{Uncertainty-Constrained A* Algorithm}
\label{alg:uc_astar}
\begin{algorithmic}[1]
\STATE \textbf{Input:} Graph $G$, start node $n_s$, goal node $n_g$
\STATE \textbf{Output:} A feasible path from $n_s$ to $n_g$
\STATE Initialize OPEN set $O \leftarrow \{n_s\}$ and CLOSED set $C \leftarrow \emptyset$
\WHILE{$O$ is not empty}
    \STATE Pick $n_{\text{best}} \in O$ such that $f(n_{\text{best}}) \le f(n),\ \forall n \in O$
    \STATE Remove $n_{\text{best}}$ from $O$ and add to $C$
    \IF{$n_{\text{best}} = n_g$}
        \STATE \textbf{EXIT}
    \ENDIF
    \FOR{all $x \in \text{Star}(n_{\text{best}})$ such that $x \notin C$}
        \STATE Validate edge $(n_{\text{best}}, x)$ using Algorithm~2
        \IF{edge is valid}
            \IF{$x \notin O$}
                \STATE add $x$ to $O$
                \STATE set backpointer of $x$ to $n_{\text{best}}$
                \STATE update $g(x)$
            \ELSIF{$g(n_{\text{best}}) + c(n_{\text{best}}, x) < g(x)$}
                \STATE update backpointer of $x$ to $n_{\text{best}}$
                \STATE update $g(x)$
            \ENDIF
        \ENDIF
    \ENDFOR
\ENDWHILE
\end{algorithmic}
\end{algorithm}

\section{THEORETICAL ANALYSIS}
\label{sec:analysis}

In this section, we provide a theoretical guarantee regarding the optimality of the proposed Uncertainty-Constrained A* planner. We formulate the problem as a search over a graph where edge validity is strictly determined by the EKF covariance bound.

\subsection{Graph Formulation}
Let $\mathcal{G} = (\mathcal{V}, \mathcal{E})$ be the connectivity graph. A node $n \in \mathcal{V}$ represents an augmented state consisting of the 
physical planning state (position) together with the associated belief 
(mean and covariance). An edge $e_{ij} \in \mathcal{E}$ exists between nodes $n_i$ and $n_j$ if and only if the transition is collision-free and satisfies the uncertainty constraint $\mathrm{Tr}(P(t)) \leq \Sigma_{\max}$ for the entire duration of the traversal.

We define the cost of an edge, denoted as $c(n_i, n_j)$, as the arc length of the trajectory generated by the carrot chase controller.

\subsection{Optimality Proof}
Standard A* search minimizes the total path cost $f(n) = g(n) + h(n)$, where $g(n)$ is the accumulated cost from the start node to $n$, and $h(n)$ is the heuristic estimate from $n$ to the goal. The algorithm is guaranteed to return the optimal solution if $h(n)$ is both \textit{admissible} and \textit{consistent}.

\textbf{Definition 1 (Heuristic Function):} We utilize the Euclidean distance as the heuristic:
\begin{equation}
    h(n) = \| \mathbf{p}_{goal} - \mathbf{p}_n \|_2,
\end{equation}
where $\mathbf{p}_{goal}$ and $\mathbf{p}_n$ is the position vector $(x, y)$ of goal node and node $n$ respectively.

\textbf{Theorem 1 (Admissibility):} The heuristic $h(n)$ is admissible.

\textit{Proof:} A heuristic is admissible if $h(n) \leq h^*(n)$ for all $n$, where $h^*(n)$ is the true minimum cost to reach the goal from $n$.
Since the robot is non-holonomic and must navigate around obstacles, the actual path length $h^*(n)$ is always greater than or equal to the straight-line Euclidean distance:
\begin{equation}
    h^*(n) \geq \| \mathbf{p}_{goal} - \mathbf{p}_n \|_2 = h(n).
\end{equation}
The uncertainty constraint removes edges from the graph but does not alter the edge costs of remaining edges. Removing edges can only increase the true cost $h^*(n)$ or make the goal unreachable ($h^*(n) = \infty$). In all cases, the inequality $h(n) \leq h^*(n)$ holds. 

\begin{figure*}[]
    \centering
    \includegraphics[
        width=\textwidth,
        height=0.5\textheight,
        keepaspectratio
    ]{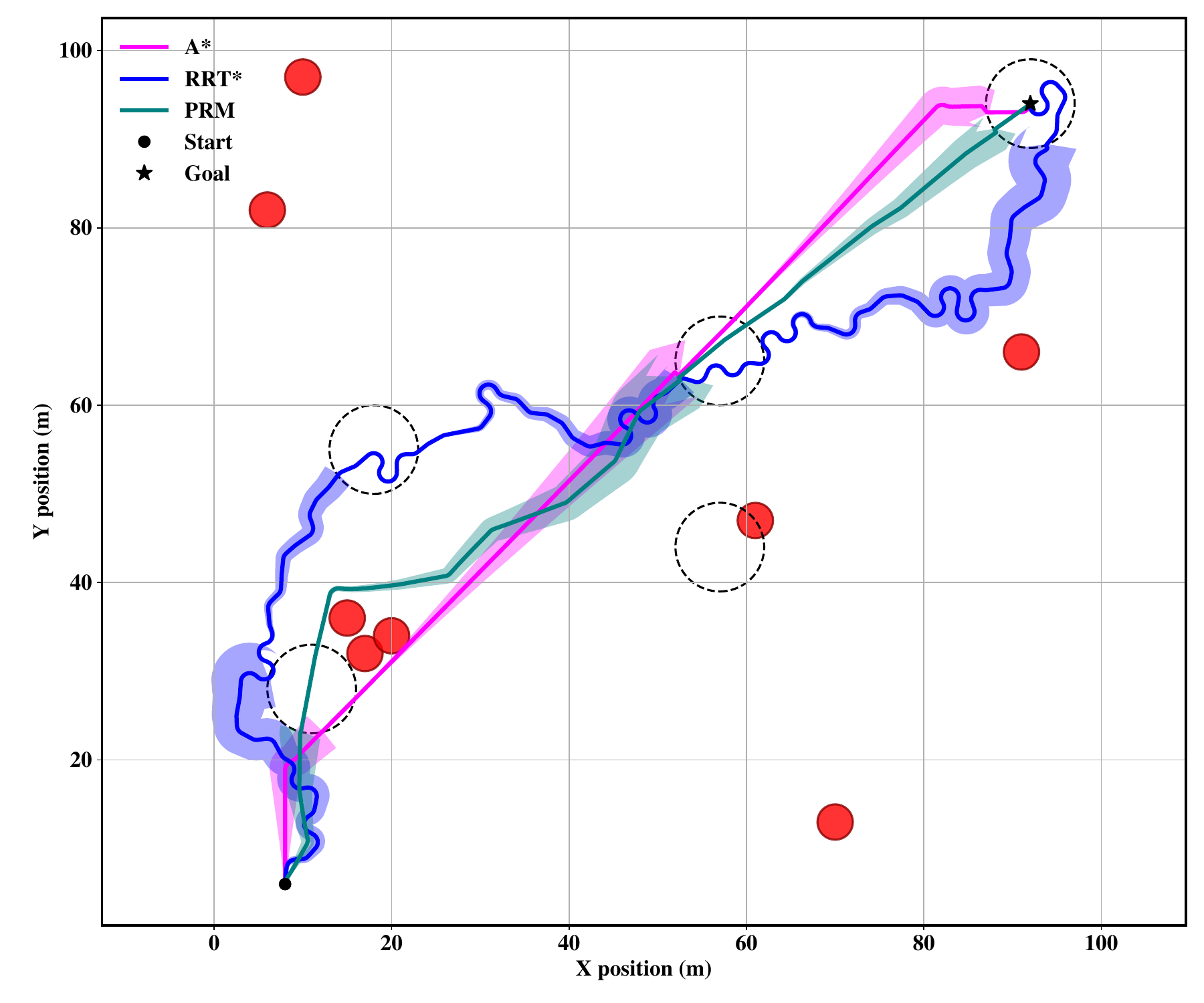}
    \caption{Trajectory generated by the uncertainty-constrained A*, RRT* and PRM planner for an environment with $1\%$ obstacle density, $1\%$ landmark density, and covariance threshold of $\Sigma_{\max} = 15$.}
    \label{fig:astar_path}
\end{figure*}

\textbf{Theorem 2 (Consistency):} The heuristic $h(n)$ is consistent.

\textit{Proof:} A heuristic is consistent (or monotonic) if for every node $n$ and its successor $n'$, the following inequality holds:
\begin{equation}
    h(n) \leq c(n, n') + h(n').
\end{equation}

Using the definition of the heuristic, we know:
\begin{equation}
    h(n) = \| \mathbf{p}_{goal} - \mathbf{p}_n \|_2 .
\end{equation}

By the triangle inequality in Euclidean space, equation 3 becomes,
\begin{equation}
    \| \mathbf{p}_{goal} - \mathbf{p}_n \|_2
    \leq
    \| \mathbf{p}_{n'} - \mathbf{p}_n \|_2
    +
    \| \mathbf{p}_{goal} - \mathbf{p}_{n'} \|_2 .
\end{equation}

The edge cost $c(n, n')$ is defined as the arc length of the feasible trajectory connecting nodes $n$ and $n'$. The length of any continuous trajectory between two points is lower bounded by the Euclidean distance between its endpoints, hence
\begin{equation}
    \| \mathbf{p}_{n'} - \mathbf{p}_n \|_2 \leq c(n, n').
\end{equation}

Substituting this bound into the previous inequality yields
\begin{equation}
    h(n) \leq c(n, n') + h(n').
\end{equation}

Therefore, the heuristic is consistent.

Since the heuristic is admissible and consistent, the Uncertainty-Constrained A* planner is guaranteed to find the path with the minimum travel distance among all feasible paths that satisfy the covariance bound $\Sigma_{\max}$.

\section{EXPERIMENTS AND RESULTS}

This section presents simulation results evaluating the proposed uncertainty-constrained planner in GPS-denied environments. The experiments demonstrate how enforcing a bound on localization uncertainty affects path feasibility, estimation accuracy, and path quality for different planners.

\subsection{Simulation Setup}
All experiments are conducted in a $100 \times 100$ two-dimensional grid space populated with static circular obstacles of radius $r_{\text{obs}} = 2.0$ and landmarks with sensing radius $r_{\text{land}} = 5.0$. The robot always starts at $(8,6)$, and the goal is fixed at $(92,94)$. 

The robot follows Dubins kinematics with constant linear velocity $v = 0.5$~m/s and minimum turning radius $R_{\min} = 1.0$. A rabbit--carrot path-following controller is used to track the planned paths. The simulation time step is $\Delta t = 0.1$~s. Further, the EKF parameters are set to
$P_o = \mathrm{diag}\big([0.01,\;0.01,\;1^\circ]\big)$ and
$R = \mathrm{diag}\big([0.25,\;0.25]\big)$.

The state covariance $P_o$ is three-dimensional as it corresponds to the full robot state $(x, y, \theta)$, whereas the measurement covariance $R$ is two-dimensional since landmarks provide only position measurements $(x, y)$. The process noise is characterized by a linear velocity standard deviation
$\sigma_v = 0.1$~m and a heading rate standard deviation
$\sigma_\phi = 2^\circ$.

Localization uncertainty is quantified using the trace of the positional covariance, $\mathrm{Tr}(P_{k})$. A constraint is enforced such that any trajectory for which
\[
\mathrm{Tr}(P_{k}) > \Sigma_{\max}
\]
at any time is rejected.

To ensure statistical robustness, all results are obtained using Monte Carlo simulations over procedurally generated environments. Obstacle density is varied from $1\%$ to $20\%$, landmark density from $1\%$ to $10\%$, and the covariance threshold is varied over $\Sigma_{\max} \in \{3.0,\,6.0,\,10.0,\,15.0\}$. 

\begin{figure}[]
    \centering
    \includegraphics[width=\columnwidth]{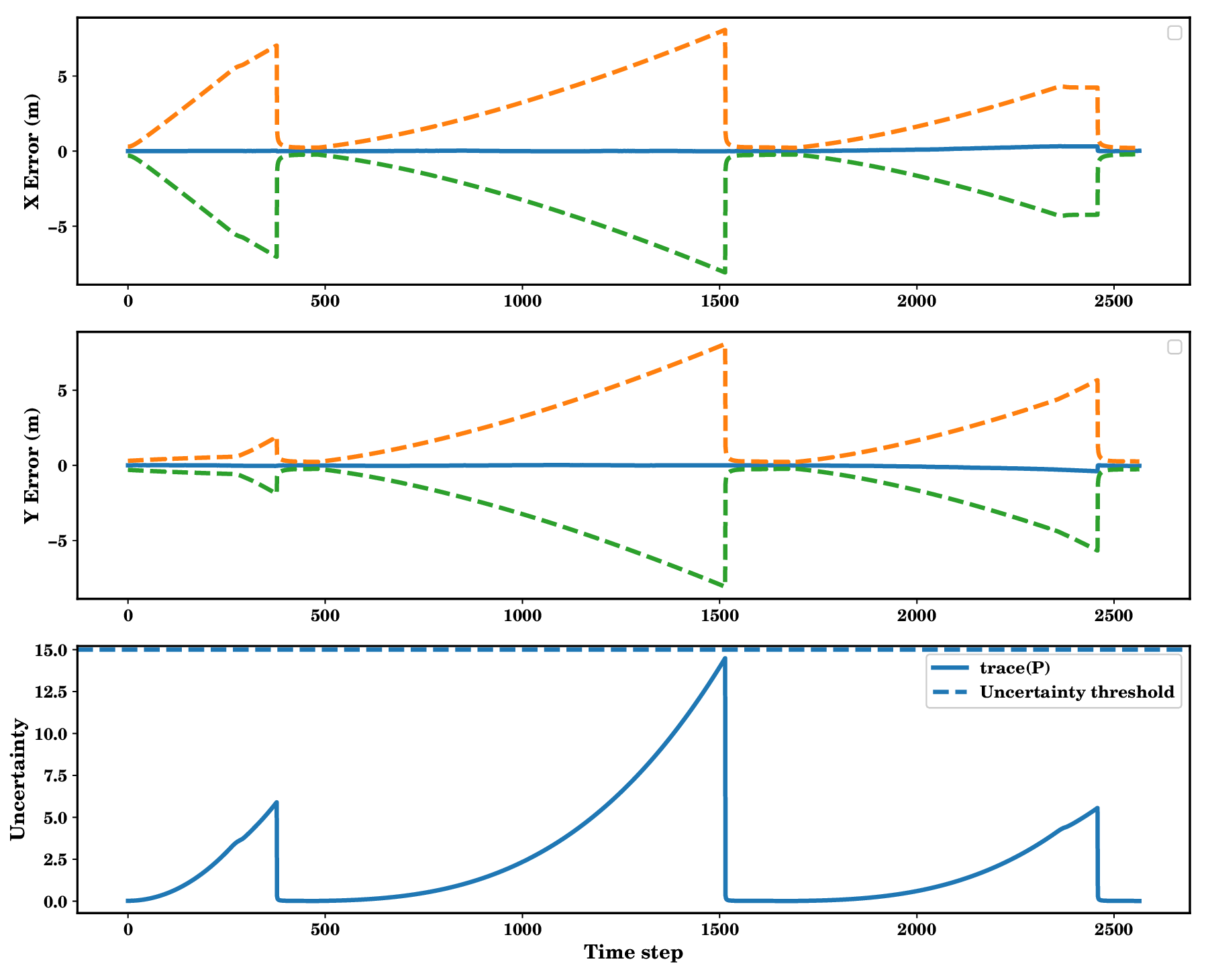}
    \caption{Tracking error and uncertainty evolution for the A* planner. Position errors remain within their $\pm 3\sigma$ bounds, and localization uncertainty stays below $\Sigma_{\max}$=15 at all times.}
    \label{fig:astar_error}
\end{figure}

\begin{figure}[]
    \centering
    \includegraphics[width=\columnwidth]{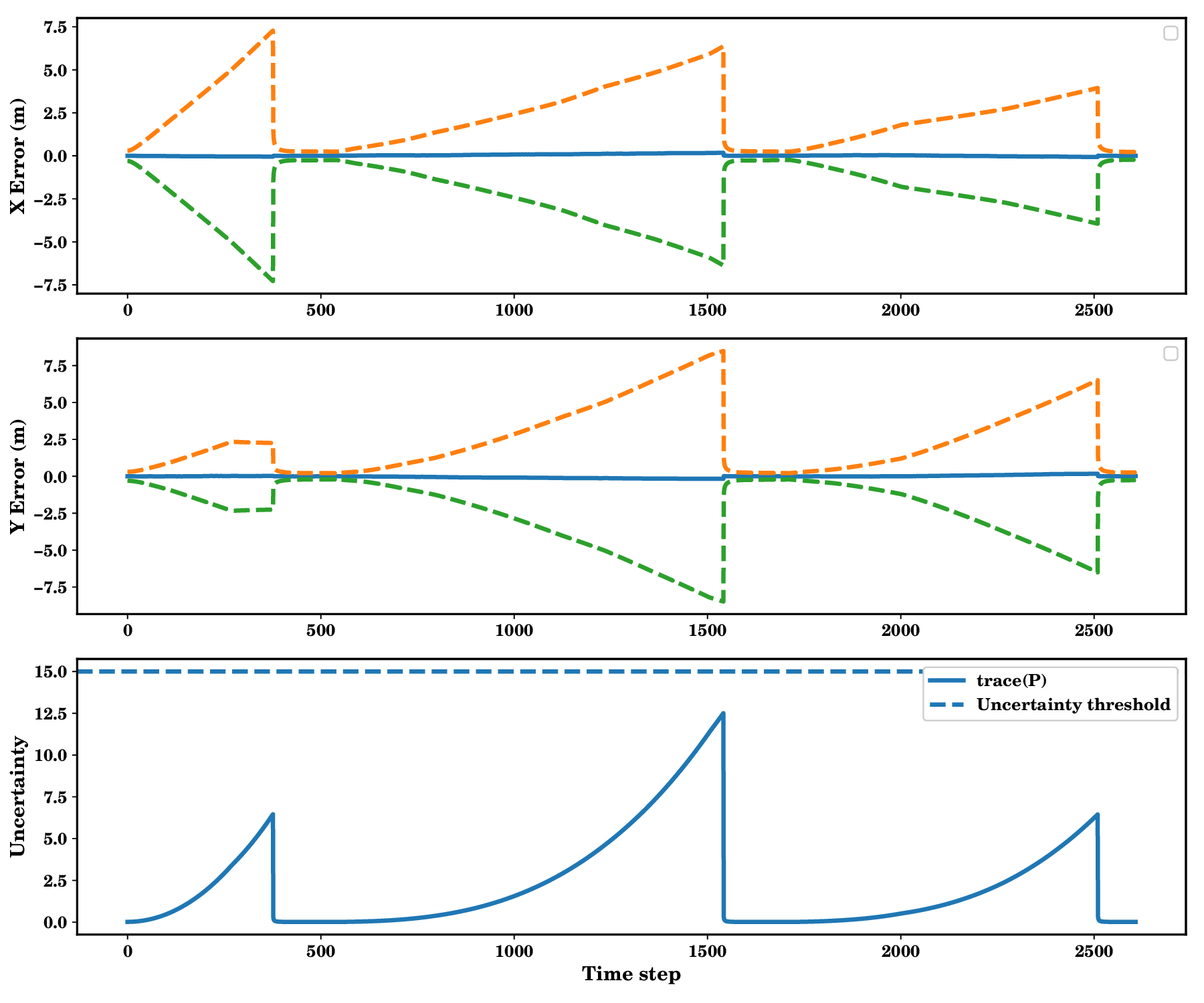}
    \caption{Tracking error and uncertainty evolution for the PRM planner. Position errors remain within their $\pm 3\sigma$ bounds, and localization  uncertainty stays below $\Sigma_{\max}$=15 at all times.}
    \label{fig:prm_error}
\end{figure}

\begin{figure}[]
    \centering
    \includegraphics[width=\columnwidth]{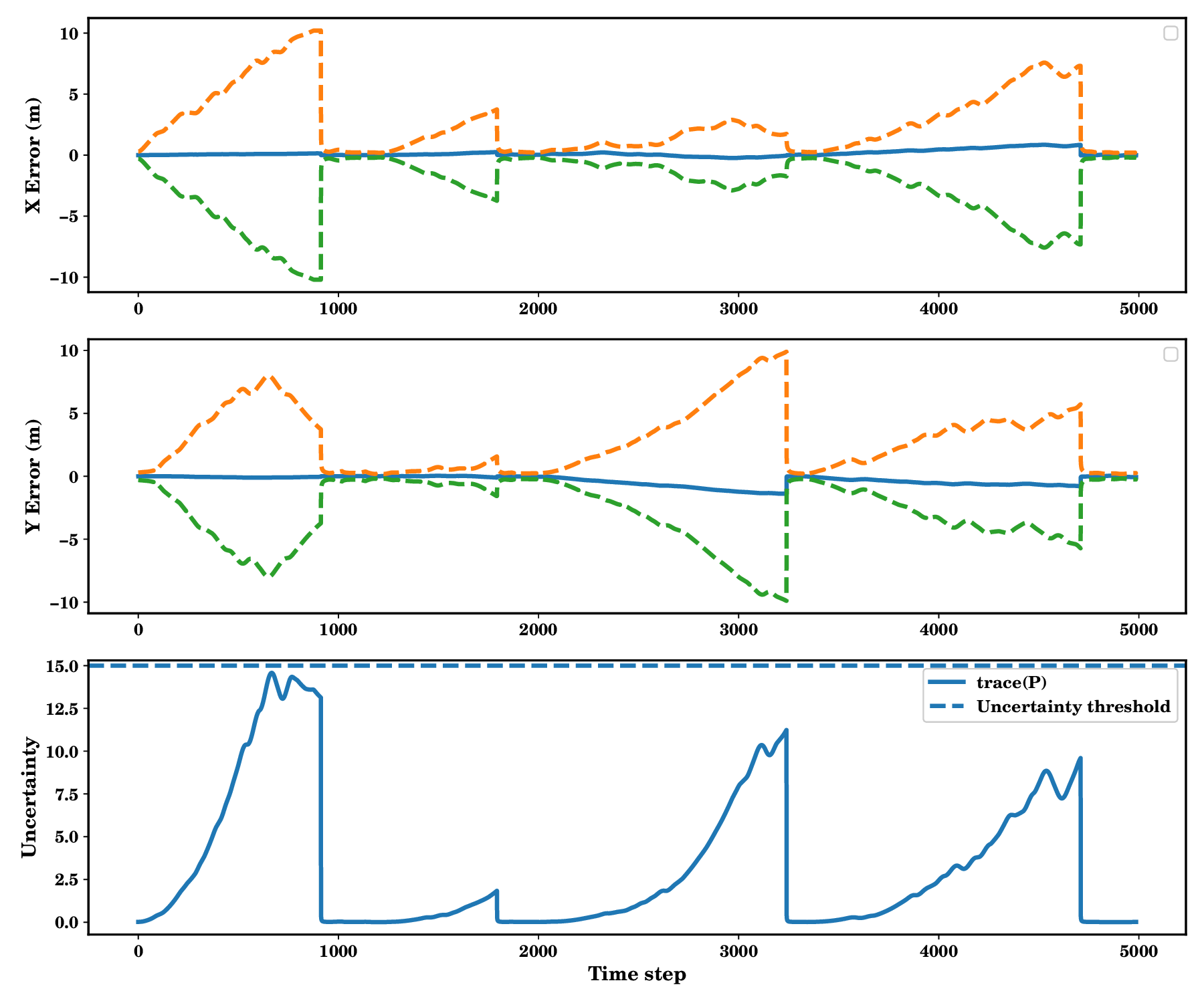}
    \caption{Tracking error and uncertainty evolution for the RRT* planner. Despite longer trajectories and requiring frequent updates, position errors remain within their $\pm 3\sigma$ bounds, and localization  uncertainty stays below $\Sigma_{\max}$=15 at all times.}
    \label{fig:rrt_error}
\end{figure}

\subsection{Planner Trajectories and Path Optimality}



The Algorithm~\ref{alg:validate} is also implemented on RRT* and PRM planners for comparison. Figure~\ref{fig:astar_path} shows representative trajectories generated by the uncertainty-constrained A*, PRM, and RRT* planners, respectively, for an environment with $1\%$ obstacle density, $1\%$ landmark density, and covariance threshold of $\Sigma_{\max} = 15$. As shown in Figure~\ref{fig:astar_path}, localization uncertainty grows during dead-reckoning segments and is reduced upon entering landmark sensing regions.

Under this configuration, A* yields the shortest uncertainty-feasible path with a total length of $128.2$~m. This result follows from its deterministic graph search and admissible heuristic, which guarantee the minimum path length among all trajectories that satisfy the uncertainty constraint.
The PRM planner generates a slightly longer trajectory of $133.0$~m. Although the resulting path satisfies the uncertainty bound, its geometry depends on the connectivity of the sampled roadmap, leading to less direct routes compared to A*. Since PRM relies on random sampling, it does not provide guarantees of finding the shortest uncertainty-feasible path.

The RRT* planner produces the longest trajectory, with a total path length of $198.3$~m. The trajectory exhibits noticeable detours caused by randomized tree expansion, which often traverses regions with poor landmark coverage before converging toward the goal. While RRT* is asymptotically optimal in unconstrained settings, the imposed uncertainty constraint significantly limits its ability to refine paths, and no finite-time optimality guarantees can be provided.



Figures~\ref{fig:astar_error}, \ref{fig:prm_error}, and \ref{fig:rrt_error} show the estimation error and uncertainty evolution for the A*, PRM, and RRT* planners, respectively. Each figure contains three plots: position error in $x$, position error in $y$, and the trace of the positional covariance.

For all planners, the position errors remain within their corresponding $\pm 3\sigma$ bounds throughout execution. This indicates that the EKF remains consistent and that estimation errors are well captured by the predicted covariance.

The uncertainty plots show that $\mathrm{Tr}(P_{k})$ increases during dead-reckoning intervals and decreases sharply upon entering landmark sensing regions. In all cases, the uncertainty remains strictly below the imposed threshold $\Sigma_{\max}$ at all times, confirming that the uncertainty constraint is satisfied throughout execution.

\begin{figure}[]
    \centering
    \includegraphics[width=\columnwidth]{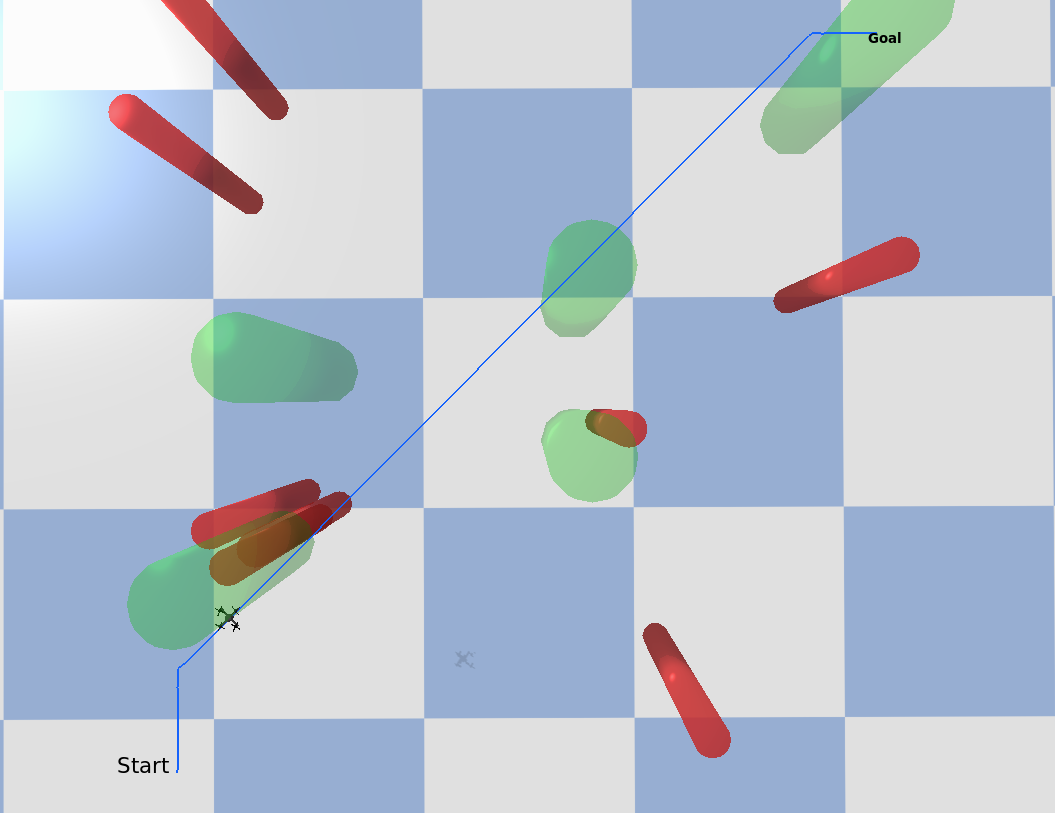}
    \caption{PyBullet quadrotor simulation showing the planned trajectory (blue), obstacles (red), and landmarks(green). The uav follows the uncertainty-constrained path while navigating through landmark regions.}
    \label{fig:pybullet_traj}
\end{figure}

\begin{figure}[]
    \centering
    \includegraphics[width=\columnwidth]{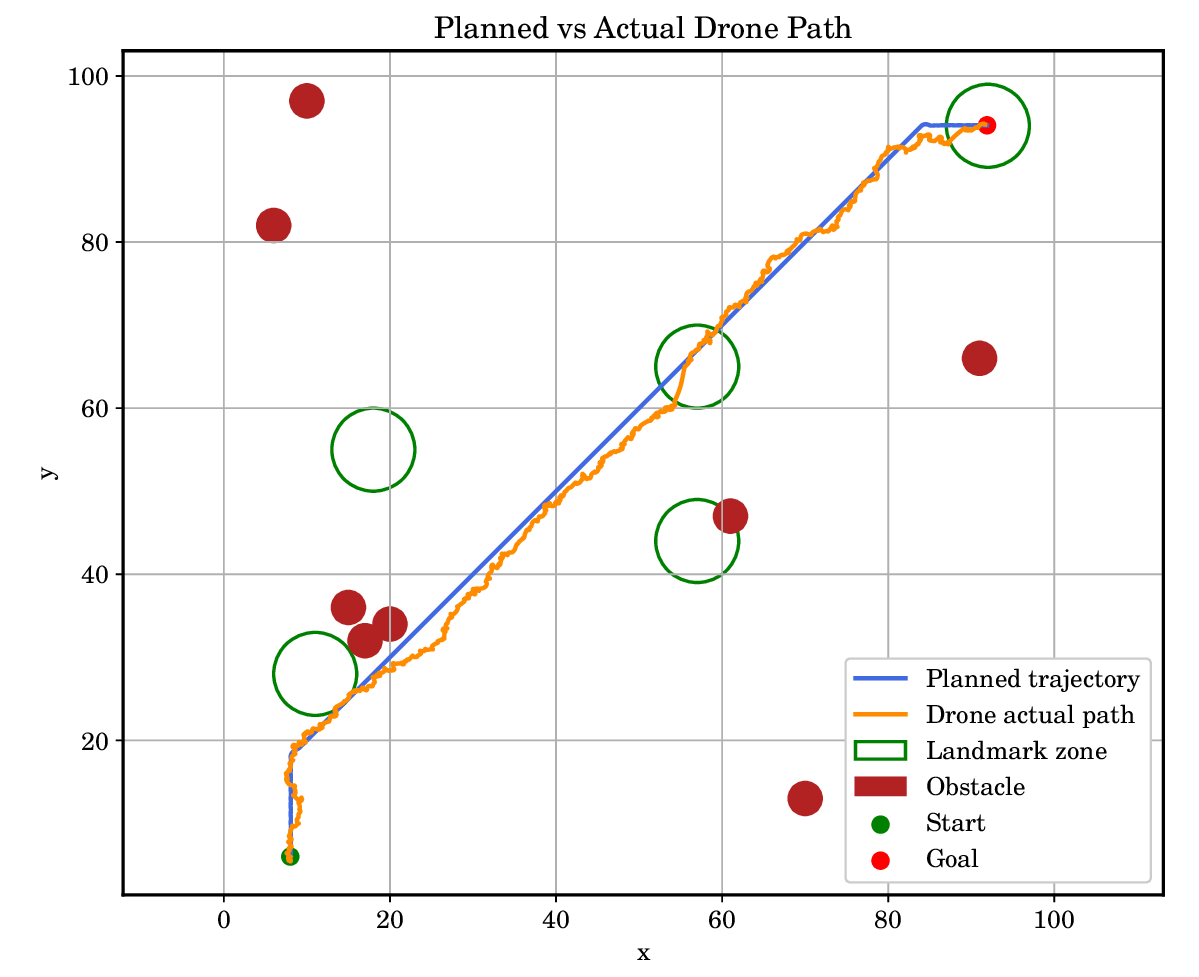}
    \caption{Tracking and estimation performance in PyBullet simulation. The executed trajectory closely follows the planned path, demonstrating bounded localization uncertainty during flight.}
    \label{fig:pybullet_tracking}
\end{figure}

To further test the method in a more realistic setting, we use a PyBullet quadrotor simulation \cite{pybullet}. The planned 2D path is converted into a fixed-height trajectory. The drone follows this path using a PID controller, while an EKF provides the estimated position based on drone model which is then used for feedback. As shown in Fig.~\ref{fig:pybullet_traj} and Fig.~\ref{fig:pybullet_tracking}, the drone successfully follows the planned path and reaches the goal. The estimated trajectory remains close to the true path, showing that localization uncertainty stays bounded. This demonstrates that the proposed method works not only in grid simulations but also in a more realistic setup.

\subsection{Impact of Landmark Density}
We evaluate the effect of landmark density by changing the number of landmark percentage while keeping obstacle density and uncertainty threshold fixed. For each case, we report the success rate, path length, and computation time averaged over successful runs from all planners. The same procedure is used to evaluate the effects of obstacle density and uncertainty threshold.

\begin{figure}[]
    \centering
    \includegraphics[width=\columnwidth]{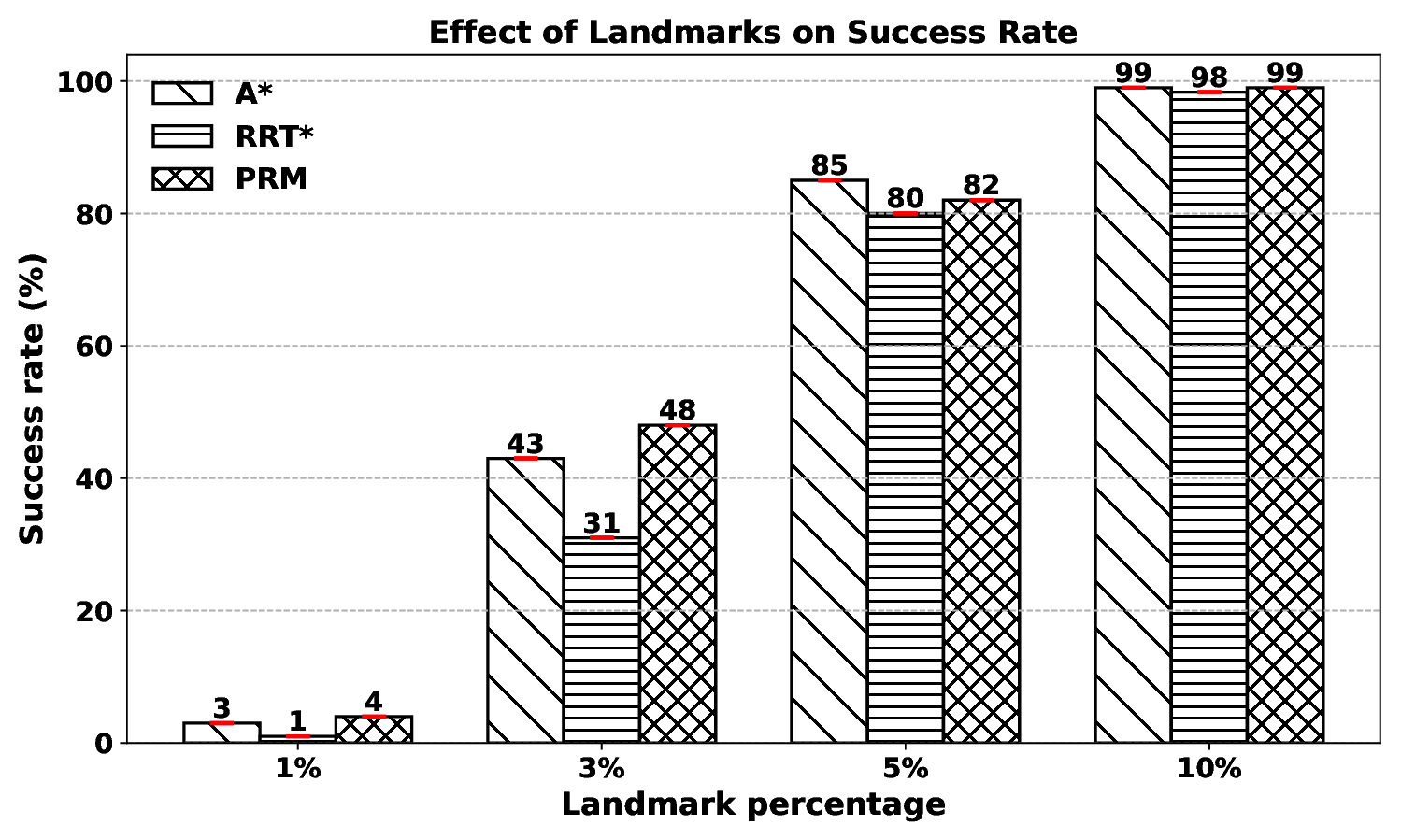}
    \caption{Effect of landmark density on success rate with obstacle percentage at 5\% and covariance threshold $\Sigma_{\max}=6$.}
    \label{fig:land1}
\end{figure}

\begin{figure}[]
    \centering
    \includegraphics[width=\columnwidth]{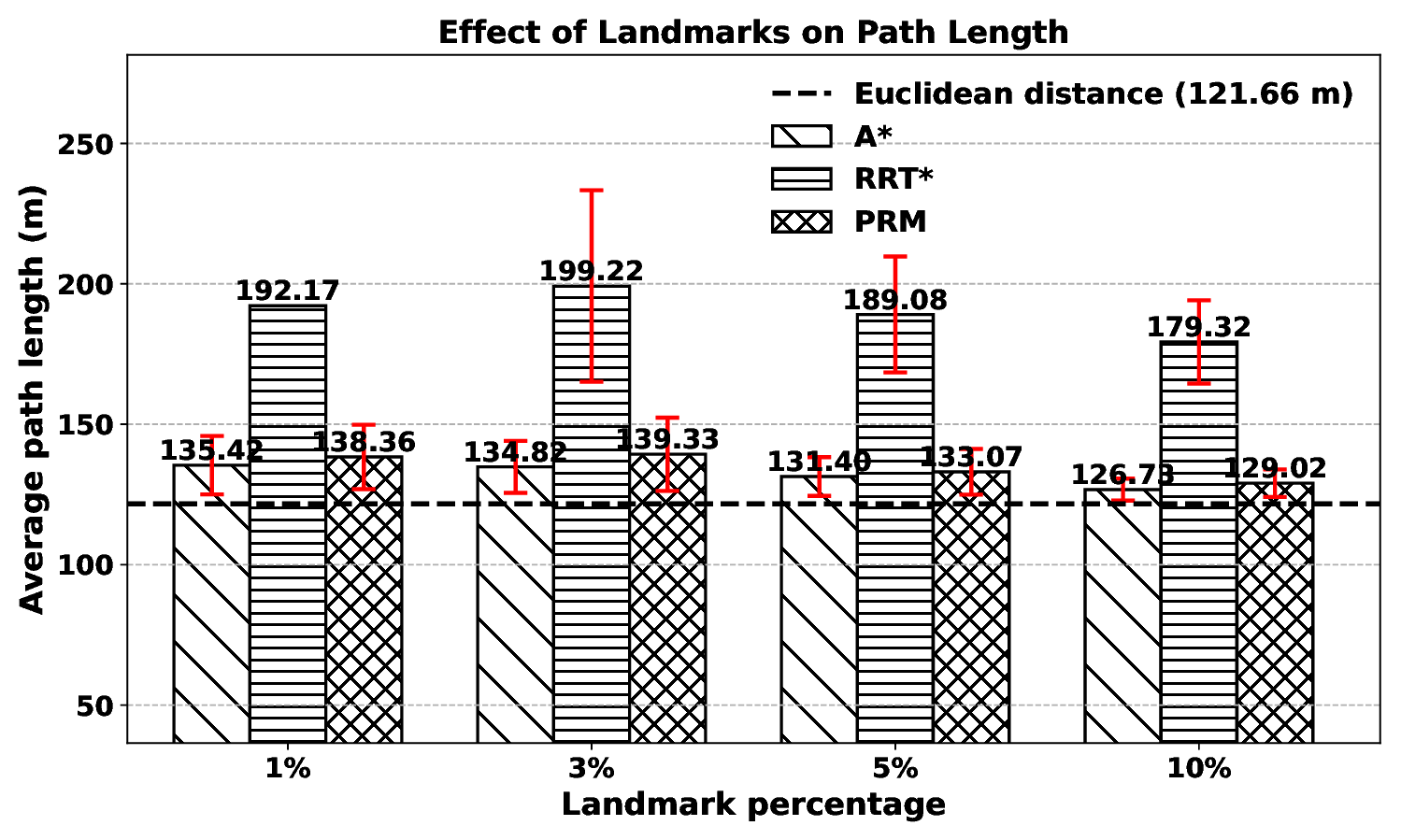}
    \caption{Effect of landmark density on mean path length with obstacle percentage at 5\% and covariance threshold $\Sigma_{\max}=6$.}
    \label{fig:land2}
\end{figure}

\begin{figure}[]
    \centering
    \includegraphics[width=\columnwidth]{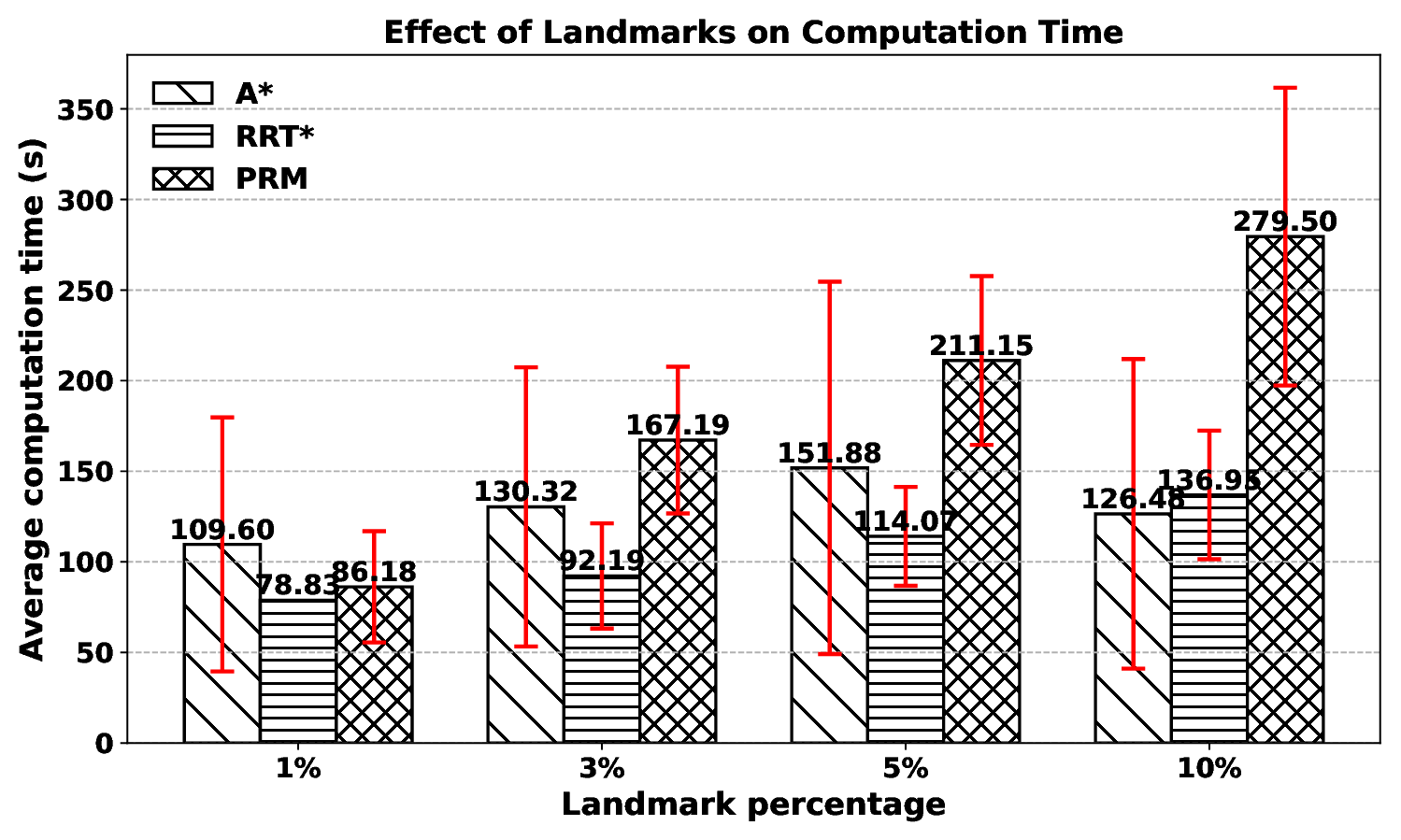}
    \caption{Effect of landmark density on computation time with obstacle percentage at 5\% and covariance threshold $\Sigma_{\max}=6$.}
    \label{fig:land3}
\end{figure}

Figures~\ref{fig:land1}–\ref{fig:land3} show how landmark density affects planner performance.
At very low landmark density (1\% and 3\%), all planners struggle to satisfy the uncertainty constraint, resulting in very low success rates. As landmark density increases to 5\%, the success rate improves steadily for all planners. A* achieves higher success than the sampling-based planners in this regime, indicating that its systematic graph search can better exploit localization observations. At the highest landmark density (10\%), all planners achieve near-perfect success rates, and the difference between methods becomes negligible.

In terms of path length (Figure~\ref{fig:land2}), A* consistently produces the shortest paths across all landmark densities. Its paths remain close to the Euclidean lower bound, reflecting the optimality guaranteed by its admissible heuristic. PRM and RRT* generate noticeably longer paths due to randomized exploration and suboptimal roadmap connectivity. Increasing landmark density reduces path length variability for all planners, but A* remains the most efficient in terms of distance traveled.

Figure~\ref{fig:land3} shows computation time. A* has the lowest computation time overall. Next is PRM planner, followed by RRT*. As landmark density increases, computation time rises for all planners.

\begin{figure}[]
    \centering
    \includegraphics[width=\columnwidth]{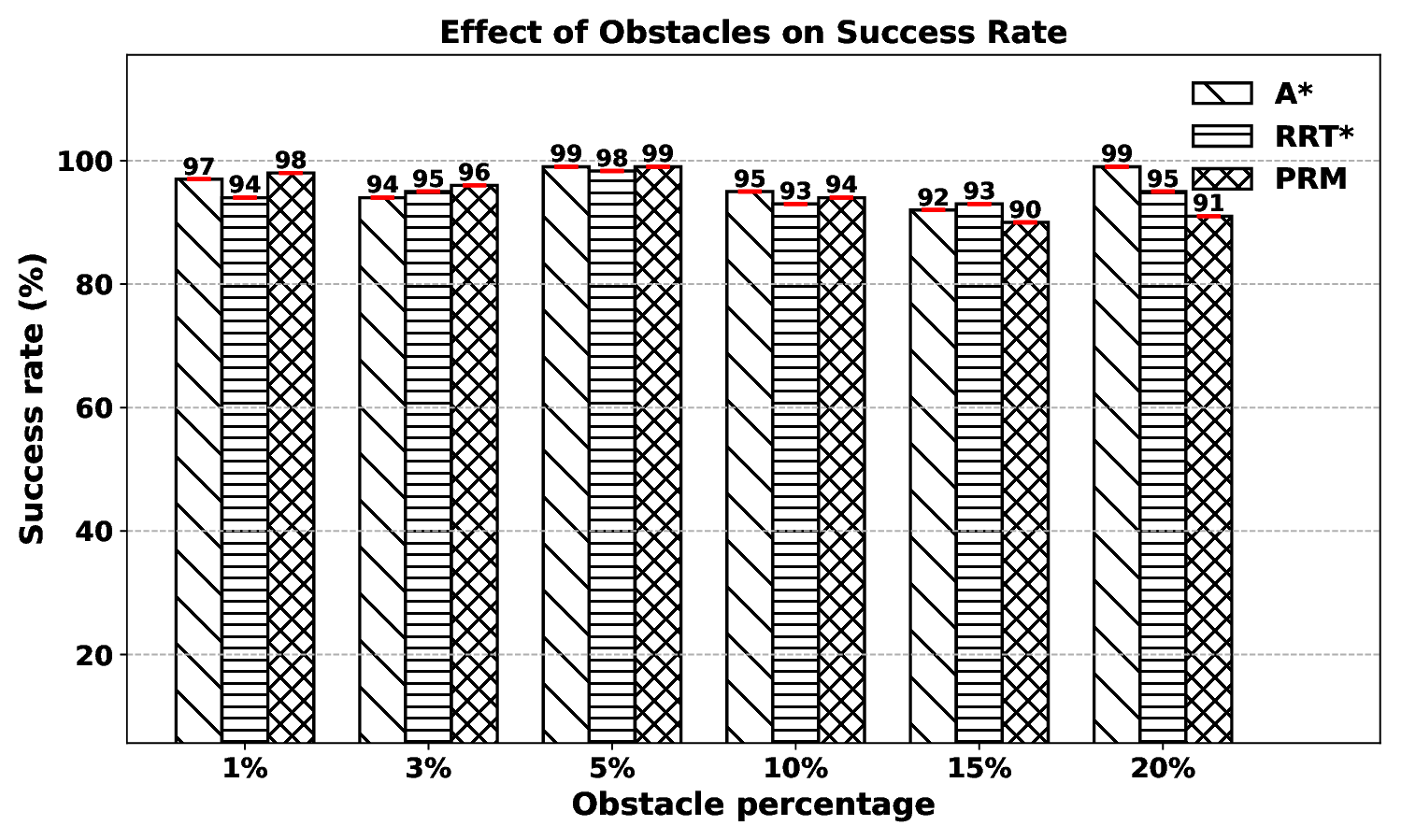}
    \caption{Effect of obstacle density on success rate with landmark percentage at 10\% and covariance threshold $\Sigma_{\max}=6$.}
    \label{fig:obs1}
\end{figure}

\begin{figure}[]
    \centering
    \includegraphics[width=\columnwidth]{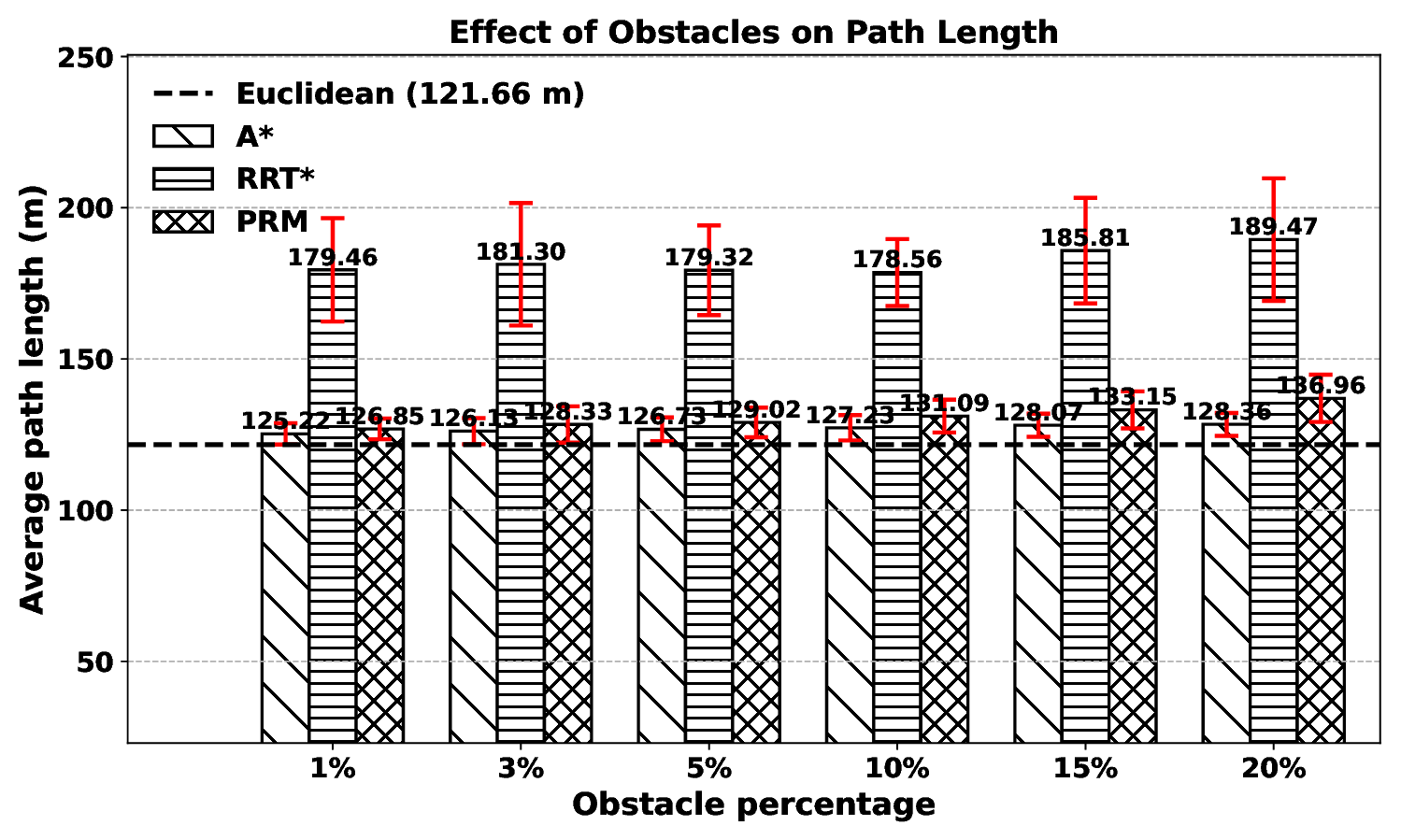}
    \caption{Effect of obstacle density on mean path length with landmark percentage at 10\% and covariance threshold $\Sigma_{\max}=6$.}
    \label{fig:obs2}
\end{figure}

\begin{figure}[]
    \centering
    \includegraphics[width=\columnwidth]{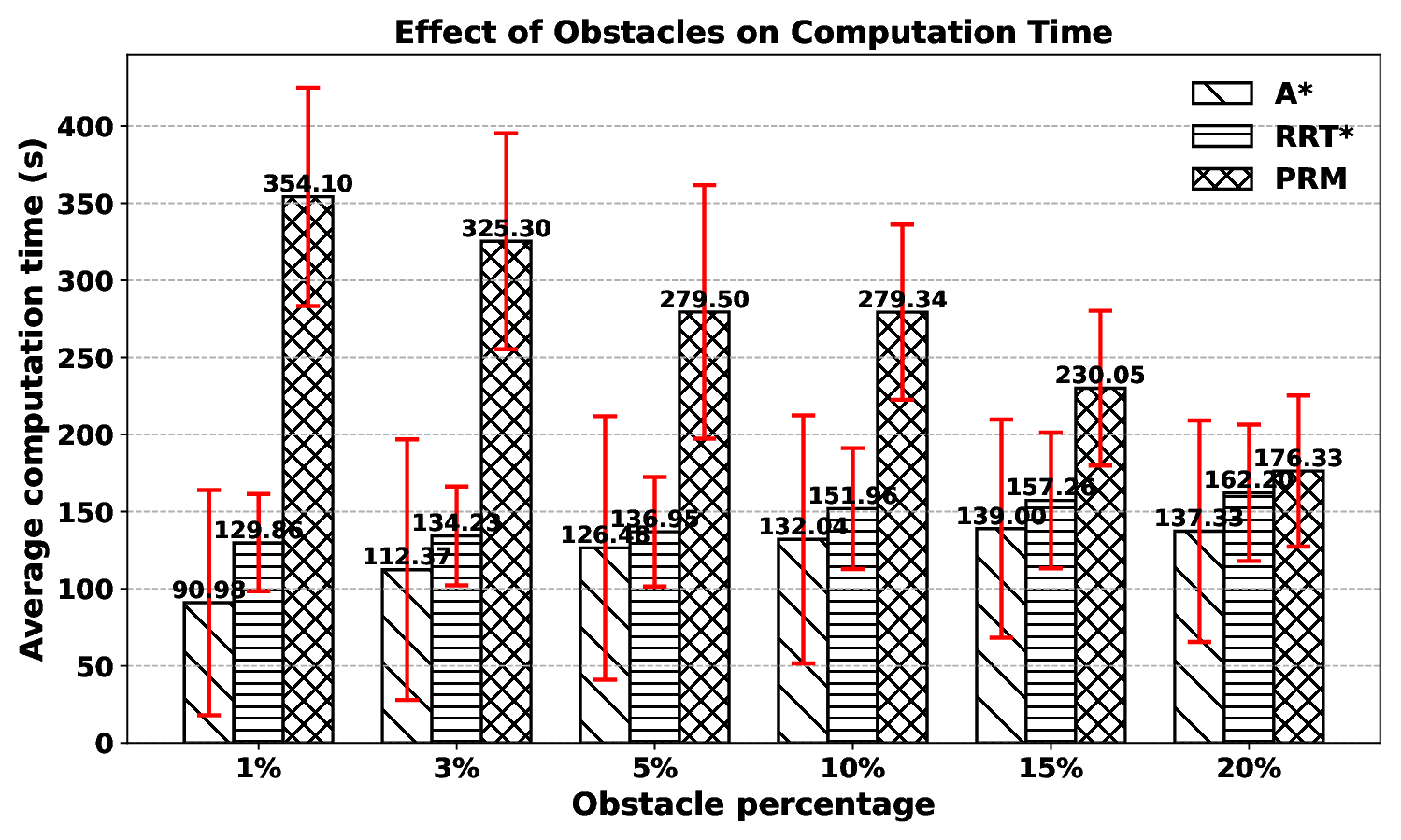}
    \caption{Effect of obstacle density on computation time with landmark percentage at 10\% and covariance threshold $\Sigma_{\max}=6$.}
    \label{fig:obs3}
\end{figure}

\subsection{Impact of Obstacle Density}

Figures~\ref{fig:obs1}–\ref{fig:obs3} illustrate the effect of obstacle density.
As shown in Figure~\ref{fig:obs1}, the success rate remains high across all obstacle densities for all planners, with only moderate variations rather than a monotonic decline. While increasing obstacle density introduces more constrained navigation, all planners maintain success rates above approximately 90\% in most cases. A* consistently achieves high success across obstacle densities, although no single planner strictly dominates in terms of feasibility under increasing obstacle clutter.

Path length results are shown in Figure~\ref{fig:obs2}. For all obstacle densities, A* consistently produces the shortest paths, demonstrating robustness of its optimality guarantees even in cluttered environments. PRM and RRT* generate longer paths due to detours caused by random sampling and limited local connectivity. As obstacle density increases, path length increases slightly for all planners, but the relative ordering remains unchanged.

Figure~\ref{fig:obs3} shows that A* has the lowest computation time at all obstacle densities, PRM next, followed by RRT* reflecting the increased number of node expansions required in cluttered environments for sampling based planners.

\subsection{Impact of Uncertainty Threshold}

\begin{figure}[]
    \centering
    \includegraphics[width=\columnwidth]{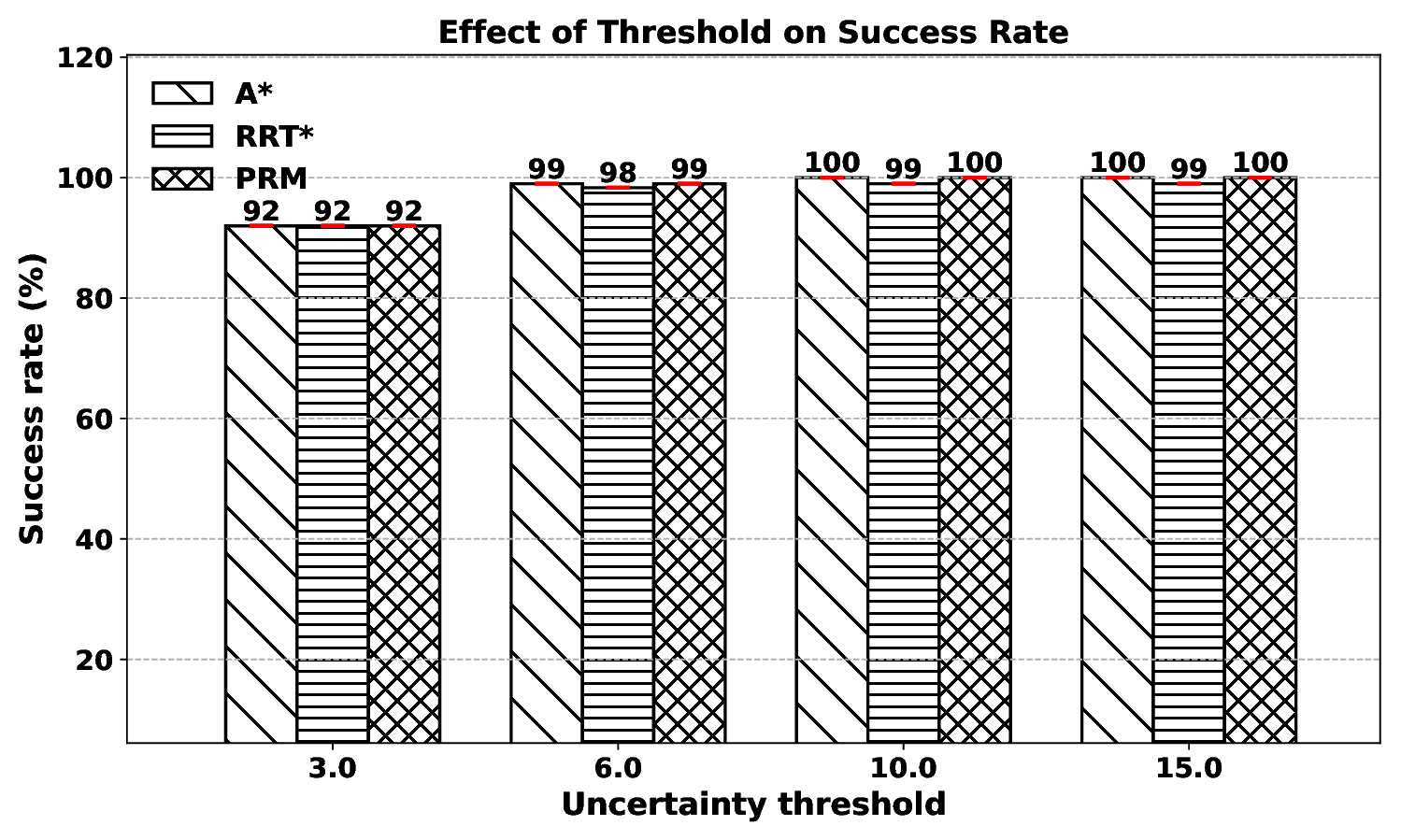}
    \caption{Effect of covariance threshold on success rate with landmark percentage at 10\% and obstacle percentage at 5\%.}
    \label{fig:th1}
\end{figure}

\begin{figure}[]
    \centering
    \includegraphics[width=\columnwidth]{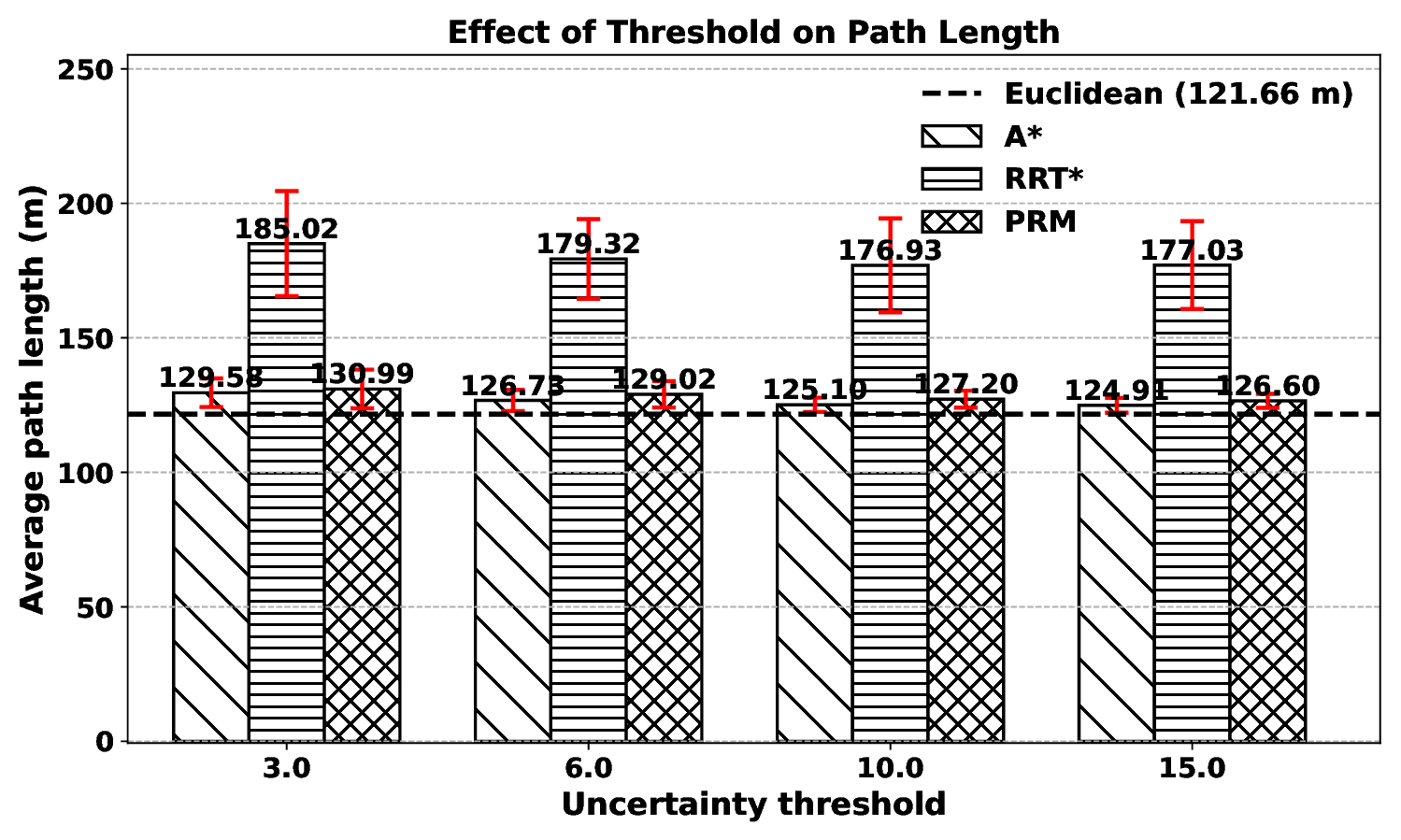}
    \caption{Effect of covariance threshold on mean path length with landmark percentage at 10\% and obstacle percentage at 5\%.}
    \label{fig:th2}
\end{figure}

\begin{figure}[]
    \centering
    \includegraphics[width=\columnwidth]{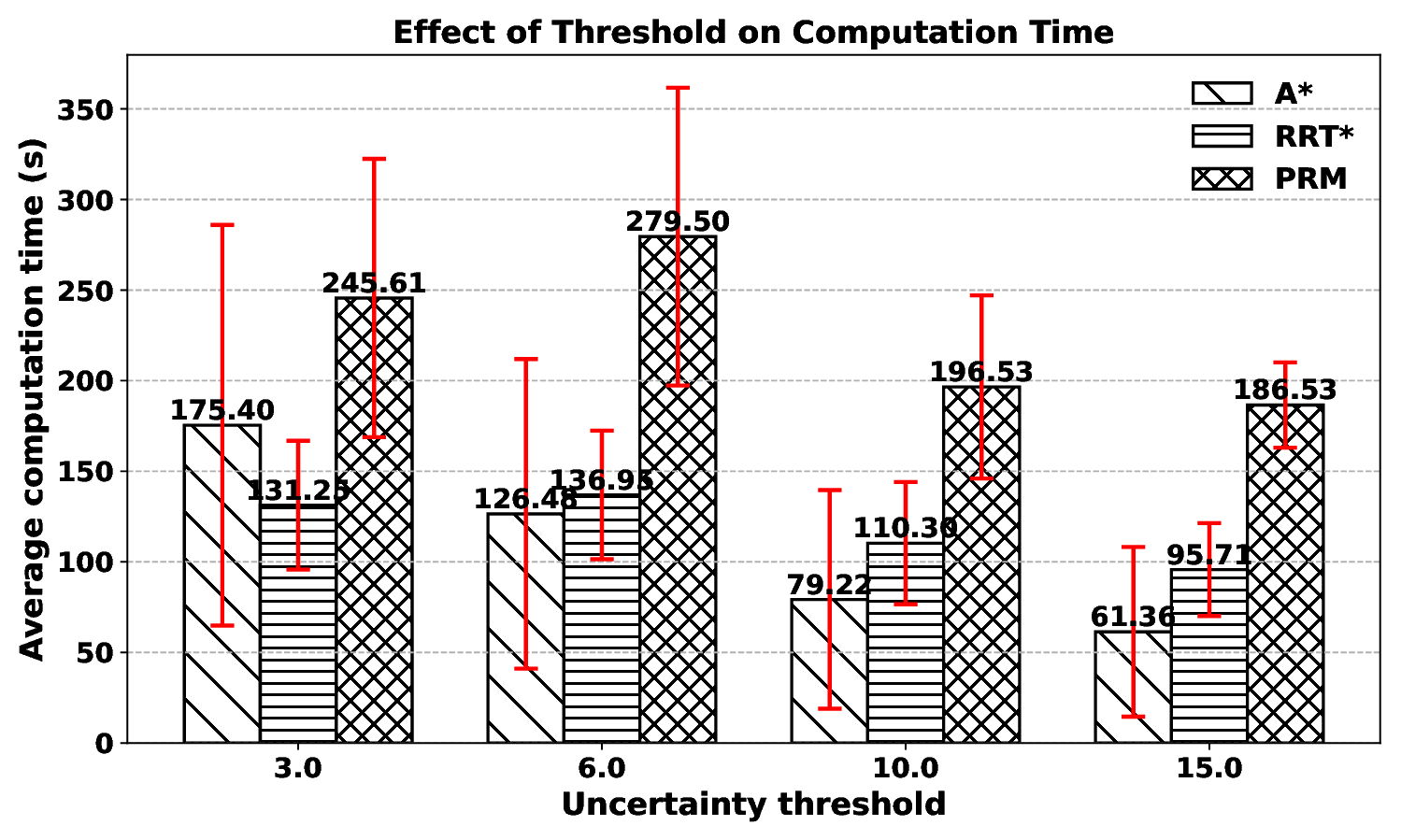}
    \caption{Effect of covariance threshold on computation time with landmark percentage at 10\% and obstacle percentage at 5\%.}
    \label{fig:th3}
\end{figure}

Figures \ref{fig:th1}–\ref{fig:th3} show the effect of the uncertainty threshold with a fixed landmark density of 10\% and obstacle density of 5\%. 

As shown in Figure \ref{fig:th1}, all planners achieve high success rates even at low thresholds, indicating sufficient landmark coverage. Increasing the uncertainty threshold further improves success rates, reaching nearly 100\% for all planners. Figure \ref{fig:th2} shows that there's only slight change in average path length when increasing the threshold. A* consistently produces the shortest paths, while RRT* and PRM generate slightly longer trajectories. Figure \ref{fig:th3} shows that computation time decreases as the uncertainty threshold increases, since fewer constraints cause replanning. A* shows the largest reduction in computation time and becomes comparable to or faster than RRT* at higher thresholds and is always faster than PRM.

\section{CONCLUSIONS}

This paper presents path planning in GPS-denied environments where localization uncertainty grows due to dead reckoning. We showed that uncertainty-constrained A* preserves its optimality guarantees and returns the shortest path among all feasible trajectories that satisfy the uncertainty constraint while also showing low computation time overall. 

Experimental results show that while geometrically optimal paths are often invalid under uncertainty constraints, our method successfully identifies feasible detours through landmark-rich areas. However, our approach assumes a known environment and known landmark locations, which may not always hold in real GPS-denied scenarios. Future work will focus on relaxing these assumptions, handling dynamic obstacles, and exploring inter-agent belief filters for multi-robot navigation under uncertainty.











\printcredits

\bibliographystyle{cas-model2-names}

\bibliography{cas-refs}



\end{document}